\documentclass{article}

\PassOptionsToPackage{numbers, compress}{natbib}

\usepackage[final]{icbinb_neurips_2020}

\usepackage[utf8]{inputenc} 
\usepackage[T1]{fontenc}    
\usepackage{hyperref}       
\usepackage{url}            
\usepackage{booktabs}       
\usepackage{amsfonts}       
\usepackage{nicefrac}       
\usepackage{microtype}      
\usepackage{cleveref}

\title{Pitfalls in Machine Learning Research:\\Reexamining the Development Cycle}

\author{%
  Stella Biderman\\
  The AI Village\\
  Booz Allen Hamilton\\
  \texttt{stellabiderman@gmail.com}
\And
  Walter J. Scheirer\\
  The AI Village\\
  University of Notre Dame\\
  \texttt{walter.scheirer@nd.edu}
}

\begin{document}
\maketitle

\begin{abstract}
    Applied machine learning research has the potential to fuel further advances in data science, but it is greatly hindered by an \textit{ad hoc} design process, poor data hygiene, and a lack of statistical rigor in model evaluation. Recently, these issues have begun to attract more attention as they have caused public and embarrassing issues in research and development. Drawing from our experience as machine learning researchers, we follow the applied machine learning process from algorithm design to data collection to model evaluation, drawing attention to common pitfalls and providing practical recommendations for improvements. At each step, case studies are introduced to highlight how these pitfalls occur in practice, and where things could be improved. 
\end{abstract}

\section{Introduction}

There is much to be excited about in the field of machine learning these days. From championship video game AI to advances in autonomous vehicles, it seems that no matter where we look we see machine learning transforming yet another domain. But while the field has been taking its victory lap for these successes, problems have surfaced from the depths of the machine learning development cycle that potentially jeopardize the entire endeavor by producing illusory experimental effects. In some cases these problems are systematic, and extend beyond the existing discussion of problems related to dataset bias. Without an adequate response to solve them, the utility of machine learning as a constructive technology is brought into question.  

Pitfalls can emerge at three critical points in the applied machine learning development cycle:
\begin{enumerate}
    \item In the design process of an algorithm, when a team is formed to solve a problem, assumptions about the problem and solution are formulated, and an algorithm is developed.
    \item  At the point of data collection for a model that will be trained using the newly developed algorithm.
    \item During evaluation, where the model's performance as a solution to the problem is assessed. 
\end{enumerate}

There are serious technical and ethical ramifications when there is a breakdown at any critical point in the development cycle. Illusory experimental effects lead to machine learning papers that not only fail to replicate, but also create serious and often difficult-to-diagnose problems for the people who try to implement the research in practice. Recent examples of this include algorithms for determining criminal tendencies \citep{wu2016automated,hashemi2020criminal} and detecting sexual orientation \citep{wang2018deep} from photos of faces, as well as commercial services that automatically determine suitable job candidates for hiring managers \citep{burke_2019}. In all of these cases, the claimed functionality is not what it appears to be due to underlying problems in experimental design, data use, and evaluation \citep{petition,aguera2018algorithms,ajunwa2019automated}.

It is important to note that we are not merely claiming that some research is morally dubious. While some of the research we discuss in this paper \textit{is} morally dubious, we see the same problems in research that isn't, including neural architecture search, the ImageNet dataset, and route-finding algorithms. Problems in the machine learning development cycle also appear when machine learning is applied to other fields of science. For example, \citet{liu2019comparison} found that out of tens of thousands of papers on deep learning published in the medical imaging literature, only a tiny fraction were methodologically sound.

The objective of this paper is to identify and illustrate common pitfalls in the machine learning development cycle. It is not to shame or embarrass particular researchers or organizations. For every example we cite in this paper, there are a dozen other examples that could have been given. Furthermore, all of the pitfalls raised in this paper are problems that have come up in our own work. The first author has withdrawn submitted papers after issues we raise were brought to her attention and has had to explain to end-users that models they developed could not be put into practice due to methodological shortcomings. The second author has worked on data-driven modeling for years, and regrets not always scrutinizing data sources and delivering evaluations that could have been more rigorous. 

While some of the issues are things that individual researchers can address in their work, many are not. Even when the problems can be addressed individually, the culture of machine learning research fails to enforce important behavioral norms. We aim to fix that.

The bulk of the rest of this paper traces the development cycle of machine learning research. We begin by discussing algorithm design, before moving on to data collection and finally model evaluation. Within each topic we raise some of the pressing problems with current machine learning practice, elucidate these problems with case studies, and then present recommendations to improve the current practice of the field. In the final section of this paper we discuss takeaways, with a focus on things the reader can 

\section{Designing the Right Algorithm}

Before one can answer the question ``Am I doing the research right?'' one must first tackle the question ``Am I doing the right research?'' Although ethical issues with machine learning research have attracted increasing attention recently, we contend that many projects suffer from fundamental design flaws that make them --- even on a purely technical level --- non-starters from the beginning.

\subsection{Problems with Algorithm Design}

\textit{Not Engaging with Stakeholders.} Modern machine learning and data science has national and even global repercussions, but researchers rarely admit the scope of the interests involved. While progress has been made in getting researchers and engineers to identify users as stakeholders, as machine learning and data science technologies become increasingly popular for public use the importance of recognizing anyone who may interact with the system as a stakeholder becomes crucial. For example, the stakeholders in FBI predictive policing algorithms~\citep{fbi-pred-pol} include every person who will set foot in the U.S. while the algorithms are in use and the stakeholders in the routing algorithms in Google Maps and Waze include the people whose neighborhoods and lives are disrupted by having drivers rerouted by their homes \citep{la-waze1,la-waze2}. Existing notions of participatory design in machine learning are not sufficient to address these sorts of situations.

\textit{Ignoring Assumptions.} Algorithms necessarily rely on oversimplified models of the world, but not enough attention is paid to the assumptions that are designed into models and the impacts that they have on the results. This is a perpetual problem in computer science, famously highlighted a decade ago in the blog post ``Falsehoods Programmers Believe About Names'' \citep{falsehoods}. Assumptions that are built into algorithms and datasets (often implicitly) can influence the results of research in ways that  undermine its credibility \citep{misgender-machines,tinyimages_retraction} or dissuade people from pursuing certain types of research \citep{politics-adv,AIV-panel}.

\textit{Lack of Oversight.} When most scientists seek to do a study that impacts humans, they are required to obtain approval by an Institutional Review Board (IRB) whose job it is to protect the rights and interests of the subjects of the study. Unfortunately, most applied machine learning and data science research is exempt from requiring IRB approval because it analyzes preexisting data. This often means that machine learning research is not reviewed by a third party until it is given to peer-reviewers, a point far too late to make meaningful changes to the design of the experiment or algorithm.

\subsection{Case study: Gender and Machine Learning}

Automated gender recognition (AGR) is a textbook example of how assumptions made in the design process can invalidate research. One example of this is how they operationalize gender without attending to the direction of causation. There are hundreds of papers that claim to be able to accurately determine people's gender based on anything from a writing sample \citep{gender-writing1,gender-writing2,gender-writing3}, to their gait \citep{gender-gait1,gender-gait2,gender-gait3}, to the texture of their skin \citep{gender-skin1,gender-skin2,gender-skin3}. Regardless of what one thinks constitutes ``gender,'' almost nobody believes that one's gender is determined by their handwriting, gait, or skin care regimen. Indeed, many of the study authors explicitly acknowledge this disconnect before proceeding to ignore it.

The typical justifications of these methodologies uses a V-shaped causation pattern: something causes gender, that same thing causes the attribute measured, and therefore we can infer things about gender from the measurement. This is invalid reasoning, but it is highly prevalent in machine learning research. When these algorithms are being used to prescriptively assign gender labels to people, something that all of the mentioned papers cite as an application of their work, it is not sufficient to invoke correlation instead of causation as a justification of the effect. None of the referenced papers employ sufficient controls to make any sort of causal claims, but every paper makes them.

AGR research also highlights how ontological assumptions can interfere with research. According to \citet{misgender-machines}, 94.8\% of AGR papers define gender in a binary fashion with no acknowledgment of the existence of transgender people. By prescriptively defining the bins that people must fall into, the algorithm is incapable of understanding humans beyond the scope of the author's conceptions. This is a very common way for researchers to imprint their own biases into their data, and best practices in fields such as medicine \citep{better-gender-med} and HCI \citep{better-gender-hci} strongly discourage it.

\subsection{Best Practices: Recommendations for Algorithm Design}

\textit{Not every problem needs to be solved via an algorithm.} Although as machine learning researchers we are inherently biased towards trying to solve problems with machine learning, many issues can be solved by simply \textit{not trying to solve them at all}. In a conversation about whether or not Twitter's automatic cropping algorithm is racially biased, Amy \citet{cropping} pointed out
\begin{quote}
    The \*easiest\* fix for that biased cropping AI? No it’s not to build another AI -- it's to give people the power to select crop boundaries when posting a photo.
\end{quote}
A similar sentiment shared by many others. Twitter took this criticism to heart, and announced changes to cropping including giving users control over how their images were cropped \citep{twitter-blog}

\textit{Democratize your notions of stakeholdership.} Any person who produces data used in a model, uses a model, or whose life is impacted by the outputs of a model is a stakeholder. This especially includes groups who are\footnote{It is typical to say ``who are historically'' here, but this is misleading and casts diversification as a solved problem which it is not.} undervalued and underrepresented in the design process.

\textit{Integrate stakeholder feedback into the default ML development cycle.} While the recent paper ``Participation is not a Design Fix for Machine Learning'' \citep{sloane2020participation} raises important points about the limits of how participatory design is often done, that does not mean that participatory design is not a crucial component of the ML development cycle. It is vital that machine learning researchers take the criticisms of their research by stakeholders seriously, making meaningful changes to their concepts and approaches rather than ``participation-washing'' their research. \citet{good-stakeholders} and \citet{sloane2020participation} provide actionable recommendations for researchers looking to improve their next study.

\textit{Subject research to ethical review.} Recent strides towards improving the ethicality of machine learning research have been made by NeurIPS's requirement of a Broader Impacts statement. While this is a good first step, ethical review needs to be taken more seriously industry-wide, and integrated earlier in the research process to prevent significant amounts of time and money being spent on fundamentally invalid or harmful research. Even when the problems with research later comes out (e.g., predicting criminality from facial structure, training language models with hate speech) that doesn't prevent the harm done when people take this research and base deployed AI algorithms on it.

\section{Collecting the Data}

The adage ``what is counted counts'' has never been more true than it is today. Troves of data are collected and stored every day, much of it to be analyzed by machine learning algorithms. How that data is collected and what precisely it counts is vital to understanding the resulting analysis. Yet far too often questions of data methodology are ignored by industry and academic researchers alike.

\subsection{Problems with Data Collection}\label{subsec:data-problems}

\textit{Statistical and Social Bias.} At this point, it should not be surprising to anyone working on machine learning to hear that datasets are heavily biased. However, bias remains a key problem facing all fields pursuing data-driven modeling. Biases tend to manifest in one of two ways: statistical and social. Statistical biases can take the form of class imbalance or repeating irregularities associated with both salient and non-salient information in data points. While recent discussion makes this seem as though it were a newly discovered problem, statistical bias in machine learning has been studied for decades~\citep{kotsiantis2006handling}. Social biases can be more subtle and tend to occur as artifacts of a data collection process or as reflections of societal biases.

\textit{Not Testing on Data Collected from the Real World.} The machine learning community has decided that widely-used benchmark datasets are the best way to evaluate an algorithm's performance. Unfortunately this can cause assessments of an algorithm's reliability in a paper to be wildly different from what happens in the real world \citep{CV-western}. The only way to evaluate an algorithm's performance in the real world is on freshly collected real world data from the population that the algorithm will actually be applied to. Machine learning can look to the field of robotics for best practices, as it has always done this and never hesitates to point out the mismatch between datasets and the real world~\citep{sunderhauf2018limits}.

\textit{The Limitations of the Big Data Paradigm.} The dominant strategy in AI product development has been to collect vast troves of data for high-capacity machine learning models. There can be no doubt that big data has led to advances in numerous applications, from photo tagging to autonomous vehicles. But this approach has its limitations, some of which are only now becoming apparent. Perhaps obvious in retrospect, ``bigness'' can be a liability. For many datasets, there are too many images for manual scrutiny or meaningful validation. Researchers have no good way to exhaustively examine each and every datum, which can lead to problematic data samples being provided to an algorithm during training. Recent work by \citet{prabhu2020large} highlights major issues with several commonly used image datasets, and how to address them in research.

\textit{Noisy Labeling.} Because we lean heavily on human annotators for supervised learning, what is inherently a noisy labeling process is prone to mistakes. Rater reliability can be assessed \citep{rodrigues_learning_2017,peterson_human_2019}, but rarely is. Nearly all datasets in common use have a single label assigned by one person to each sample. This does not give us any indication about the correctness or difficulty of the sample, and can lead to problems down the line during training.  The labeling process is also prone to malicious attack. It has been shown that it is possible to change labels in such a way that provides a favorable outcome to an attacker (e.g., a backdoor in a trained model) \citep{chen2017targeted}. The problem is exacerbated by big data: if a small number of labels change in a sea of millions of samples, does anybody notice? Those changes may be consequential to model outcomes. 

\subsection{Case Study: The Tiny Images Dataset}

A very recent example that reflects almost all of the above problems is the retraction \citep{tinyimages_retraction} of the Tiny Images dataset \citep{torralba200880}, which had been used for object recognition research. This dataset contains very small images ($32 \times 32$ pixels) for over 50,000 different noun categories, and was meant to develop visual recognition capabilities that match the ``remarkable tolerance of the human visual system.'' As of this writing, the paper describing the dataset has been cited over 1,700 times, and numerous algorithms have been developed using it as source data. The retraction was prompted by an investigation by \citet{prabhu2020large}, who noted that Tiny Images used several categories for images labeled with racial and misogynistic slurs. Under closer scrutiny, they found that the dataset also contained non-consensual pornography such as up-skirt photographs and imagery degrading to various marginalized groups. 

The problems that were exposed in Tiny Images map directly to the problems we have singled out related to bias, big data, and labeling. Tiny Images most obviously suffered from a case of social bias in its racist and misogynistic categories. These were unfortunate reflections of Internet culture, which were unavoidable in the collection strategy used by the creators of the dataset: an automated data procedure that relied on nouns from WordNet \citep{miller1998wordnet}. This saga also exemplifies the limitations of the big data paradigm. According to the retraction issued by Torralba et al. ``The dataset is too large (80 million images) and the images are so small (32 x 32 pixels) that it can be difficult for people to visually recognize its content.'' Thus, it is argued that the very advantage of big data is rendered moot by the presence of even a small number of problematic data samples. Because it is not possible to find all of the problematic instances in a dataset, according to Torralba et al., the only recourse is to take a dataset out of service if problems are found. And when it comes to labels, Tiny Images contained  arbitrary label assignments that reflected accepted and derogatory racial categories in WordNet. Moreover, the crawling process relied on tags from the web, not multiple assignments from a collection of annotators. Without a consensus judgment, this means the accuracy of the labels remains unclear for many of the images. Finally, there is a temporal aspect to the labels: assigned labels can evolve over time as social and cultural norms change. 

\subsection{Best Practices: Recommendations for Data Collections}

\textit{Hypothesis Driven Data Collection.} Instead of simply hoovering up data and then trying to use it for whatever arbitrary application that comes down the line, a better approach would be to design a collection with a hypothesis in mind. There is no reasonable expectation that any question can be answered by using a generic pool of data, no matter how large. Across the natural sciences, experimental data collection is tightly coupled with a specific hypothesis that is formulated before work begins. The same should be true of experiments in machine learning. 

\textit{Auditing and Documenting Datasets.} While we acknowledge that it is impossible to go exhaustively through today's machine learning datasets by hand, there are still some sensible strategies for auditing that can be used. Given that most dataset come from sites where users can upload their own content, there are fairly obvious problems to look for in a targeted way: hate speech, profanity, and pornography. Less obvious catches can be made by auditing the sources of the data. Are celebrity news sites overly represented in a dataset for photo captioning? That may lead to racial bias in operation \citep{Scheirer_Bulletin_2020}. Thus better heterogeneity in sourcing is needed. In software engineering, a set of tests that is not comprehensive, but still useful to reveal failure modes is known as \textit{smoke testing}. This idea transfers nicely to dataset auditing, where feasible checks for the above items and others can made in a reasonable amount of time. From the point of view of documentation, the ``Datasheets for Datasets'' \citep{datasheets} framework is something that people are beginning to use and which would benefit the field if adopted as a standard \citep{datasheet1,datasheet2,datasheet3}.

\textit{Quantify Annotator Uncertainty.} When it comes to the quantification of annotator uncertainty, the recommendation here is to quantify aleatoric uncertainty \citep{kiureghian_aleatory_2009,kendall_what_2017}. This is the  uncertainty estimated and attempted to be removed when aggregating data from different annotators. With a quantitative value of uncertainty for a datapoint, a decision can be made to use or not use that point,  or perhaps weight its influence appropriately, which has been shown to be effective~\cite{plank2014learning}. The process can include an assessment of how familiar annotators are with the domain, as well the social and cultural norms associated with the type of data being annotated.  

\textit{Dataset Revision Process.} The above problems mean that at some point in a dataset's life-cycle, it will need to revised. A revision of a dataset can remove problematic information, document what has been removed from the previous version, and provide an explanation for why the revision was necessary. This may not completely remove all problems from a dataset, especially in a big data context, but it is a way to address specific problems as they are raised. Revision control systems for data should be developed to ease this process. Importantly, a standard for dataset revision should be defined and adopted by the community. ``Datasheets for Datasets'' \citep{datasheets} is a framework that has attracted attention for this, but researchers and practitioners have been slow to put it into practice.

\section{Evaluating the Model}

The field of machine learning is founded upon dataset-based evaluation. On the one hand, using standard datasets provide a common basis for comparison across different algorithms, as well as sufficient data for self-contained evaluations. On the other hand, because datasets are self-contained worlds they often misrepresent algorithm performance and can result in misleading findings if datasets are too heavily relied on. Exacerbating these problems are intentional or unintentional misuses of learning algorithms. Here we make specific recommendations on how evaluation can be improved informed by common practices in other experimental fields and previous observations from the machine learning literature that have gone unheeded. These recommendations are made for artificial neural networks, but in some cases can apply to other learning algorithms as well.

\subsection{Problems with Model Evaluation}\label{subsec:problem-eval}

\textit{Lack of Rigorous Statistical Evaluation}. Statistically verifying the results of a scientific paper is essential to its publication and the credibility of its results. Despite the attention to rigor in the algorithm development process, many of the most popular and successful machine learning models do not apply statistically rigorous techniques in evaluation. This can take many forms, including failing to report interval estimates for results, failing to analyze variance and random seed effects, and failing to properly control for covariates such as training methodology.

A machine learning model is usually considered successful if it can exceed state-of-the-art performance on standard datasets, but researchers rarely pay attention to the statistical significance of their results. Making judgments based on a small number of data points with little attention to the significance of their findings leads researchers to incorrectly over- or under-value work. In the worst cases, the lack of significance testing leads to a failure to notice that methodologies do not beat basic null models for the task.

A largely unacknowledged problem in neural network experiments is the lack of $k$-fold testing by varying the value of the random seed during training. Rarely do we find a paper that reports error based on this form of evaluation. It is well known that different random initializations of the elementary parameters can lead to drastically different results \citep{pinto2009high}. For datasets with fixed training, validation, and testing partitions, this presents a dilemma. An honest experimenter may get lucky or unlucky, depending on the choice of the seed. Thus the reported results may not reflect what happens on average for a series of training runs. A dishonest experimenter may attempt to mine seeds for an extended period of time, looking for a favorable starting place that leads to good results on the test set.

\textit{Failing to Compare to Null Models.} Not all accuracies are created equal. On some tasks, e.g., malware detection, getting to 90\% accuracy is trivial while on others, e.g., predicting the outbreak of war, it would be world-changing. While this fact is well-known to researchers, it is not sufficiently respected by them.

A recent example of this is ``Criminality from Face'' research, where deep learning was reported to be able to determine whether or not somebody is a criminal based on a photo of their face \citep{wu2016automated,hashemi2020criminal}. \citet{bowyer2020criminality} demonstrated that accurate results can be achieved for this task because of the organization of the datasets used: mugshot photos from government data sources are labeled ``criminal'' and ordinary public photos crawled from the web are labeled ``non-criminal.'' The authors found that the results were completely explained by dataset classification. In the now classic paper an ``Unbiased Look at Dataset Bias,'' \citet{torralba2011unbiased} warned about this very problem. That paper has largely been remembered for kicking off the study of individual biases within computer vision datasets, but it made a more important observation about datasets as self-contained worlds: they possess a certain visual style at a global-level, which can easily be learned. If an experimenter is labeling the datasets to suit their needs, either intentionally or unintentionally, this observation can be exploited. 

\textit{Lack of 3rd Party Evaluation.} The academic peer review process is meant to verify the scientific integrity of work, which includes reproducibility. Earlier in this paper, we noted that the vast majority of published deep learning models for medical image analysis cannot be independently verified \citep{liu2019comparison}. It is now standard practice by corporate research labs to not publish code or data with their papers in the interest of protecting intellectual property. This means that it is impossible to verify the claims made without re-implementing the work described in the paper from scratch, which is sometimes impossible if corresponding configurations are not made available to the public, or if details have been omitted. Even when code is available, replications of machine learning research often fail \citep{replication-raff}.

Aggravating this problem has been the dramatic increase in  papers submitted to machine learning-oriented publication venues. CVPR alone went from 2,123 papers in 2015 to 6,424 in 2020. Reviewers, who are overburdened even with the average load of 5-6 papers, do not have time to work with any available code under these conditions.

On the commercial side, companies are under no obligation to submit their products to peer review. In other industries, it is common practice to seek 3rd party evaluation of a product for quality or standards certification. This has rarely been the case in the AI industry. When blackbox machine learning products are released without 3rd party evaluation, problems that might otherwise have been identified in testing can emerge in operation. In one example, Microsoft released a blackbox web app for photo captioning that reproduced racial stereotypes which are prevalent in computer vision datasets composed of celebrity photos \citep{Scheirer_Bulletin_2020}. In another case,  Twitter's blackbox photo cropping algorithm demonstrated racial bias, and led someone to start a public experiment on the platform \citep{twitter}. Presently, we only see change when there is large public outcry, which isn't a sustainable strategy. 

\subsection{Case Study: Neural Architecture Search and Randomly Wired Neural Networks}

Neural architecture search (NAS) is a field of deep learning that tries to optimize the neural network architecture and find networks that produce better results when trained. Unfortunately neural architecture search by and large does not work, and many NAS researchers seem to have not noticed this fact due to poor statistical practice.

The landmark paper by \citet{rand-wire} points out that that NAS papers typically compare against each other with no external reference points or null models to compare to. To address this gap, they train neural networks whose computational graphs are generated randomly. To decrease the influence of any bias or knowledge that the authors have on the null models, they decide to generate graphs using standard random graph generators from social network analysis. These random graph generators have the additional benefit of having never been applied to NAS before. The surprising result was that some of the random network generators achieved ``state-of-the-art'' performance.

Unfortunately, the neural architecture search community does not seem to have learned much from --- or even understood --- \citet{rand-wire}. At the time of writing \citet{rand-wire} had 144 citations. Of those, more than $75\%$ of the papers cite it as an example of neural architecture search being effective, with more than $25\%$ \textit{explicitly describing it as a state-of-the-art methodology}. Despite the fact that hundreds of NAS papers have been published since \citet{rand-wire}, subsequent research has continued to validate the fact that basic null models based on random search remain the ``state-of-the-art'' for NAS \citep{nas-random-search,nas-local-search,nas-remarkable}. This term itself is a bit of a misnomer though, as any methodology that fails to outperform random networks is at a very basic level failing to do optimization at all.

\citet{nas-remarkable} look into the problems with NAS research deeper, replicating eight recent NAS systems on five image datasets. They find evidence that all of the problems identified in \cref{subsec:problem-eval} are widespread in NAS:

\begin{quote}
    \textbf{Lack of Statistical Rigor:} Poor statistical rigor has allowed for confounding factors to cast doubt on or invalidate results on widely used search spaces. In particular, they find that the Differentiable Architecture Search (DARTS) space is not capable of producing real NAS innovation as the exact training set-up and choice of hyperparameters cause a significantly large variation in results than architecture improvements.
    
    \textbf{Lack of Null Models:} None of the papers in question rigorously compare to adequate null models, and all NAS methods examined either do not improve over null models or do not significantly do so.
    
    \textbf{Lack of Replicability:} While the eight NAS systems they replicated were selected because they had open source code, the authors note that this is not common. Additionally, they discuss that many NAS papers are non-comparable due to using very different search spaces.
\end{quote}

\subsection{Best Practices for Model Evaluation} 

\textit{Statistically Validate Results.} A better option than reporting results with just a point estimate is to report the average and standard error of the estimator. This can be done by varying the random seed, subsampling the test dataset, or by hypothesis testing. While some of these techniques are catching on, cultural norms around meaningful levels of statistic rigor in published research are needed.

\textit{Train a Dataset Classifier as a Control Model.} In order to determine whether or not a dataset or data source is being learned instead of the intended function, our recommendation is to perform a control experiment that swaps the target labels in the training set for dataset labels. This will lead to the creation of a dataset classifier at training time if a dataset is being learned. This process is especially important in cases where multiple datasets are being used in the training set. If this detector is able to correctly identify the source of new instances, it is likely learning the statistical differences between datasets rather than the property of interest.

\textit{Conduct Third Party Evaluation.} Authors of machine learning papers deserve another set of eyes on their work. Conferences and journals should give reviewers access to the exact code, data and configuration for the experiments described in papers introducing new algorithms. Area chairs and area editors should expect reviewers to run and tinker with the code, and referee reports should include an analysis of reviewer experiments focusing on replication and sensitivity to free parameters (including a change of data). This will inevitably slow down the rate of publication, but improve the quality of published work (and perhaps ease the burden on the conference reviewing process). It is very much in-line with recent calls for machine learning to join the Slow Science movement~\citep{bengio-slow}. With respect to commercial products, an organization similar to Underwriters Laboratories (\url{https://www.ul.com/}) should be established to certify the safety and correctness of machine learning products. There is already some precedent for this in computer security \citep{UL}.

\section{Takeaways for Doing Better Data Science}

The central challenge for methodologists is to convince practitioners that their suggestions for improvement are worth pursuing. Resistance to changing methodologies that are perceived as working is quite reasonable -- as one reviewer of this paper put it, ``Hypothesis-driven data collection: this sounds good on paper, but what does it mean in practice... Even if cost were no issue, ML has made enormous progress using standard benchmark datasets, such as ImageNet, to allow for objective comparison of results from different labs using different methods. Are the authors suggesting that the field abandon this approach?''

While we agree that the methodologies commonplace in applied machine learning research has produced great success, that doesn't mean that we should not continue to strive to do better. To this point, we have sought to not only criticize current scientific practices but also to elevate work that exemplifies or instructs research to a higher standard. We hope that this, together with our analysis of prominent failure cases at each stage in the development process, inspires researchers to hold themselves and each other to higher standards.

If there is one thing the reader takes away from this paper, we hope it is that there are concrete steps that individuals bring home with them to improve their methodological practices. This includes thinking about whether or not a problem needs to be solved via an algorithm, applying more rigorous standards of statistical analysis, auditing and documenting datasets, and comparing results to null models. These recommendations can be implemented on a project-by-project basis without significant external support. Additionally, although many of our suggestions require larger scale changes than one person can accomplish alone, you can encourage the development of the necessary norms and cultural attitudes by asking these questions in peer review, posing them to project leads, and making your personal commitment to them known. Even if people do not yield to your questions, raising them and normalizing them as a point of discussion is an important first step.

\section*{Acknowledgments}

We would like to thank the AI Village for invaluable discussion of the themes of this paper and feedback on the manuscript.

\bibliographystyle{plainnat}
\bibliography{data_pseudoscience}

\begin{thebibliography}{59}
\providecommand{\natexlab}[1]{#1}
\providecommand{\url}[1]{\texttt{#1}}
\expandafter\ifx\csname urlstyle\endcsname\relax
  \providecommand{\doi}[1]{doi: #1}\else
  \providecommand{\doi}{doi: \begingroup \urlstyle{rm}\Url}\fi

\bibitem[Afifi(2019)]{gender-skin3}
Mahmoud Afifi.
\newblock 11k hands: gender recognition and biometric identification using a
  large dataset of hand images.
\newblock \emph{Multimedia Tools and Applications}, 78\penalty0 (15):\penalty0
  20835--20854, 2019.

\bibitem[Agrawal and Davis(2020)]{twitter-blog}
Parag Agrawal and Dantley Davis.
\newblock Transparency around image cropping and changes to come.
\newblock \emph{Twitter Blog}, 2020.

\bibitem[Ag{\"u}era~y Arcas et~al.(2018)Ag{\"u}era~y Arcas, Todorov, and
  Mitchell]{aguera2018algorithms}
Blaise Ag{\"u}era~y Arcas, Alexander Todorov, and Margaret Mitchell.
\newblock Do algorithms reveal sexual orientation or just expose our
  stereotypes?
\newblock \emph{medium.com}, 2018.

\bibitem[Ajunwa(2019)]{ajunwa2019automated}
Ifeoma Ajunwa.
\newblock Automated employment discrimination.
\newblock \emph{Available at SSRN 3437631}, 2019.

\bibitem[Albert et~al.(2020)Albert, Penney, Schneier, and
  Siva~Kumar]{politics-adv}
Kendra Albert, Jon Penney, Bruce Schneier, and Ram~Shankar Siva~Kumar.
\newblock Politics of adversarial machine learning.
\newblock In \emph{Towards Trustworthy ML: Rethinking Security and Privacy for
  ML Workshop, Eighth International Conference on Learning Representations
  (ICLR)}, 2020.

\bibitem[Argamon et~al.(2009)Argamon, Koppel, Pennebaker, and
  Schler]{gender-writing1}
Shlomo Argamon, Moshe Koppel, James~W Pennebaker, and Jonathan Schler.
\newblock Automatically profiling the author of an anonymous text.
\newblock \emph{Communications of the ACM}, 52\penalty0 (2):\penalty0 119--123,
  2009.

\bibitem[Bauer et~al.(2017)Bauer, Braimoh, Scheim, and
  Dharma]{better-gender-med}
Greta~R Bauer, Jessica Braimoh, Ayden~I Scheim, and Christoffer Dharma.
\newblock Transgender-inclusive measures of sex/gender for population surveys:
  Mixed-methods evaluation and recommendations.
\newblock \emph{PloS one}, 12\penalty0 (5):\penalty0 e0178043, 2017.

\bibitem[Bengio(2020)]{bengio-slow}
Yoshua Bengio.
\newblock Time to rethink the publication process in machine learning.
\newblock
  \url{https://yoshuabengio.org/2020/02/26/time-to-rethink-the-publication-process-in-machine-learning/},
  2020.
\newblock Accessed: 2020-9-20.

\bibitem[Bhatt et~al.(2020)Bhatt, Andrus, Weller, and Xiang]{good-stakeholders}
Umang Bhatt, McKane Andrus, Adrian Weller, and Alice Xiang.
\newblock Machine learning explainability for external stakeholders.
\newblock \emph{arXiv preprint arXiv:2007.05408}, 2020.

\bibitem[Biderman et~al.(2020)Biderman, Anandkumar, Caliskan, D'Ignazio, and
  Kumar]{AIV-panel}
Stella Biderman, Anima Anandkumar, Aylin Caliskan, Catherine D'Ignazio, and Ram
  Shankar~Siva Kumar.
\newblock Ai ethics and bias panel.
\newblock The AI Village at DEF CON, 2020.
\newblock URL \url{https://www.youtube.com/watch?v=7zswHvHR9cA}.

\bibitem[Bowyer et~al.(2020)Bowyer, King, and Scheirer]{bowyer2020criminality}
Kevin~W Bowyer, Michael King, and Walter Scheirer.
\newblock The criminality from face illusion.
\newblock \emph{arXiv preprint arXiv:2006.03895}, 2020.

\bibitem[Burke(2019)]{burke_2019}
L.~Burke.
\newblock Your interview with {AI}.
\newblock \emph{Inside Higher Ed}, 2019.

\bibitem[Chen et~al.(2017)Chen, Liu, Li, Lu, and Song]{chen2017targeted}
Xinyun Chen, Chang Liu, Bo~Li, Kimberly Lu, and Dawn Song.
\newblock Targeted backdoor attacks on deep learning systems using data
  poisoning.
\newblock \emph{arXiv preprint arXiv:1712.05526}, 2017.

\bibitem[{Coalition for Critical Technology}(2020)]{petition}
{Coalition for Critical Technology}.
\newblock Abolish the \#techtoprisonpipeline.
\newblock
  \url{https://medium.com/@CoalitionForCriticalTechnology/abolish-the-techtoprisonpipeline-9b5b14366b16},
  2020.
\newblock Accessed: 2020-9-20.

\bibitem[Costa-juss{\`a} et~al.(2020)Costa-juss{\`a}, Creus, Domingo,
  Dom{\'\i}nguez, Escobar, L{\'o}pez, Garcia, and Geleta]{datasheet2}
Marta~R Costa-juss{\`a}, Roger Creus, Oriol Domingo, Albert Dom{\'\i}nguez,
  Miquel Escobar, Cayetana L{\'o}pez, Marina Garcia, and Margarita Geleta.
\newblock Mt-adapted datasheets for datasets: Template and repository.
\newblock \emph{arXiv preprint arXiv:2005.13156}, 2020.

\bibitem[Das and Chakrabarty(2015)]{gender-gait2}
Deepjoy Das and Alok Chakrabarty.
\newblock Human gait based gender identification system using hidden markov
  model and support vector machines.
\newblock In \emph{International Conference on Computing, Communication \&
  Automation}, pages 268--272. IEEE, 2015.

\bibitem[de~Vries et~al.(2019)de~Vries, Misra, Wang, and van~der
  Maaten]{CV-western}
Terrance de~Vries, Ishan Misra, Changhan Wang, and Laurens van~der Maaten.
\newblock Does object recognition work for everyone?
\newblock In \emph{Proceedings of the IEEE Conference on Computer Vision and
  Pattern Recognition Workshops}, pages 52--59, 2019.

\bibitem[Do et~al.(2020)Do, Nguyen, and Kim]{gender-gait3}
Trung~Dung Do, Van~Huan Nguyen, and Hakil Kim.
\newblock Real-time and robust multiple-view gender classification using gait
  features in video surveillance.
\newblock \emph{Pattern Analysis and Applications}, 23\penalty0 (1):\penalty0
  399--413, 2020.

\bibitem[Eidam(2016)]{fbi-pred-pol}
Eyragon Eidam.
\newblock The role of data analytics in predictive policing.
\newblock \emph{Government Technology}, 2016.

\bibitem[Gattal et~al.(2020)Gattal, Djeddi, Bensefia, and
  Ennaji]{gender-writing3}
Abdeljalil Gattal, Chawki Djeddi, Ameur Bensefia, and Abdellatif Ennaji.
\newblock Handwriting based gender classification using cold and hinge
  features.
\newblock In \emph{International Conference on Image and Signal Processing},
  pages 233--242. Springer, 2020.

\bibitem[Gebru et~al.(2018)Gebru, Morgenstern, Vecchione, Vaughan, Wallach,
  Daum{\'e}~III, and Crawford]{datasheets}
Timnit Gebru, Jamie Morgenstern, Briana Vecchione, Jennifer~Wortman Vaughan,
  Hanna Wallach, Hal Daum{\'e}~III, and Kate Crawford.
\newblock Datasheets for datasets.
\newblock \emph{FAT ML}, 2018.

\bibitem[Hashemi and Hall(2020)]{hashemi2020criminal}
Mahdi Hashemi and Margeret Hall.
\newblock Criminal tendency detection from facial images and the gender bias
  effect.
\newblock \emph{Journal of Big Data}, 7\penalty0 (1):\penalty0 1--16, 2020.

\bibitem[Kendall and Gal(2017)]{kendall_what_2017}
Alex Kendall and Yarin Gal.
\newblock What {Uncertainties} {Do} {We} {Need} in {Bayesian} {Deep} {Learning}
  for {Computer} {Vision}?
\newblock In \emph{NeurIPS}, 2017.

\bibitem[Keyes(2018)]{misgender-machines}
Os~Keyes.
\newblock The misgendering machines: Trans/hci implications of automatic gender
  recognition.
\newblock \emph{Proceedings of the ACM on Human-Computer Interaction},
  2\penalty0 (CSCW):\penalty0 1--22, 2018.

\bibitem[Kiureghian and Ditlevsen(2009)]{kiureghian_aleatory_2009}
Armen~Der Kiureghian and Ove Ditlevsen.
\newblock Aleatory or epistemic? {D}oes it matter?
\newblock \emph{Structural Safety}, 31\penalty0 (2):\penalty0 105--112, March
  2009.

\bibitem[Kotsiantis et~al.(2006)Kotsiantis, Kanellopoulos, Pintelas,
  et~al.]{kotsiantis2006handling}
Sotiris Kotsiantis, Dimitris Kanellopoulos, Panayiotis Pintelas, et~al.
\newblock Handling imbalanced datasets: A review.
\newblock \emph{GESTS International Transactions on Computer Science and
  Engineering}, 30\penalty0 (1):\penalty0 25--36, 2006.

\bibitem[Kumar et~al.(2019)Kumar, Gupta, Sharma, and Kumar~Saroj]{gender-skin2}
Deepak Kumar, Rajat Gupta, Ashirwad Sharma, and Sushil Kumar~Saroj.
\newblock Gender classification using skin patterns.
\newblock In \emph{Proceedings of 2nd International Conference on Advanced
  Computing and Software Engineering (ICACSE)}, 2019.

\bibitem[Li and Talwalkar(2020)]{nas-random-search}
Liam Li and Ameet Talwalkar.
\newblock Random search and reproducibility for neural architecture search.
\newblock In \emph{Uncertainty in Artificial Intelligence}, pages 367--377.
  PMLR, 2020.

\bibitem[Littman(2019)]{la-waze2}
Johnathan Littman.
\newblock Waze hijacked l.a. in the name of convenience. can anyone put the
  genie back in the bottle?
\newblock \emph{Los Angeles Magazine}, 2019.

\bibitem[Liu et~al.(2019)Liu, Faes, Kale, Wagner, Fu, Bruynseels, Mahendiran,
  Moraes, Shamdas, Kern, et~al.]{liu2019comparison}
Xiaoxuan Liu, Livia Faes, Aditya~U Kale, Siegfried~K Wagner, Dun~Jack Fu, Alice
  Bruynseels, Thushika Mahendiran, Gabriella Moraes, Mohith Shamdas, Christoph
  Kern, et~al.
\newblock A comparison of deep learning performance against health-care
  professionals in detecting diseases from medical imaging: a systematic review
  and meta-analysis.
\newblock \emph{The Lancet Digital Health}, 1\penalty0 (6):\penalty0
  e271--e297, 2019.

\bibitem[Lopez(2018)]{la-waze1}
Steve Lopez.
\newblock Column: How waze and google maps turned an encino neighborhood into a
  speedway.
\newblock \emph{Los Angeles Times}, 2018.

\bibitem[McKenzie(2010)]{falsehoods}
Patrick McKenzie.
\newblock Falsehoods programmers believe about names.
\newblock \emph{Kalzumeus Software}, 2010.

\bibitem[Miller(1998)]{miller1998wordnet}
George~A Miller.
\newblock \emph{WordNet: An electronic lexical database}.
\newblock MIT press, 1998.

\bibitem[Ottelander et~al.(2020)Ottelander, Dushatskiy, Virgolin, and
  Bosman]{nas-remarkable}
T~Den Ottelander, Arkadiy Dushatskiy, Marco Virgolin, and Peter~AN Bosman.
\newblock Local search is a remarkably strong baseline for neural architecture
  search.
\newblock \emph{arXiv preprint arXiv:2004.08996}, 2020.

\bibitem[Peterson et~al.(2019)Peterson, Battleday, Griffiths, and
  Russakovsky]{peterson_human_2019}
Joshua~C Peterson, Ruairidh~M Battleday, Thomas~L Griffiths, and Olga
  Russakovsky.
\newblock Human uncertainty makes classification more robust.
\newblock In \emph{Proceedings of the IEEE International Conference on Computer
  Vision}, pages 9617--9626, 2019.

\bibitem[Pinto et~al.(2009)Pinto, Doukhan, DiCarlo, and Cox]{pinto2009high}
Nicolas Pinto, David Doukhan, James~J DiCarlo, and David~D Cox.
\newblock A high-throughput screening approach to discovering good forms of
  biologically inspired visual representation.
\newblock \emph{PLoS Comput Biol}, 5\penalty0 (11):\penalty0 e1000579, 2009.

\bibitem[Plank et~al.(2014)Plank, Hovy, and S{\o}gaard]{plank2014learning}
Barbara Plank, Dirk Hovy, and Anders S{\o}gaard.
\newblock Learning part-of-speech taggers with inter-annotator agreement loss.
\newblock In \emph{Proceedings of the 14th Conference of the European Chapter
  of the Association for Computational Linguistics}, pages 742--751, 2014.

\bibitem[Prabhu(2020)]{twitter}
Vinay Prabhu.
\newblock
  \url{https://twitter.com/vinayprabhu/status/1307460502017028096?s=20}, 2020.
\newblock Accessed: 2020-9-20.

\bibitem[Prabhu and Birhane(2020)]{prabhu2020large}
Vinay~Uday Prabhu and Abeba Birhane.
\newblock Large image datasets: A pyrrhic win for computer vision?
\newblock \emph{arXiv preprint arXiv:2006.16923}, 2020.

\bibitem[Raff(2019)]{replication-raff}
Edward Raff.
\newblock A step toward quantifying independently reproducible machine learning
  research.
\newblock In \emph{Advances in Neural Information Processing Systems}, pages
  5485--5495, 2019.

\bibitem[Rodrigues et~al.(2017)Rodrigues, Mariana, Ribeiro, and
  Pereira]{rodrigues_learning_2017}
Filipe Rodrigues, Lourenco Mariana, Bernardete Ribeiro, and Francisco~C.
  Pereira.
\newblock Learning supervised topic models for classification and regression
  from crowds.
\newblock \emph{IEEE T-PAMI}, 39\penalty0 (12):\penalty0 2409 -- 2422, 5 2017.

\bibitem[Scheirer(2020)]{Scheirer_Bulletin_2020}
W.~Scheirer.
\newblock How to make ai less racist.
\newblock \emph{Bulletin of the Atomic Scientists}, 2020.

\bibitem[Seck et~al.(2018)Seck, Dahmane, Duthon, and Loosli]{datasheet1}
Isma{\"\i}la Seck, Khouloud Dahmane, Pierre Duthon, and Ga{\"e}lle Loosli.
\newblock Baselines and a datasheet for the cerema awp dataset.
\newblock \emph{arXiv preprint arXiv:1806.04016}, 2018.

\bibitem[Sloane et~al.(2020)Sloane, Moss, Awomolo, and
  Forlano]{sloane2020participation}
Mona Sloane, Emanuel Moss, Olaitan Awomolo, and Laura Forlano.
\newblock Participation is not a design fix for machine learning.
\newblock \emph{arXiv preprint arXiv:2007.02423}, 2020.

\bibitem[Spiel et~al.(2019)Spiel, Haimson, and Lottridge]{better-gender-hci}
Katta Spiel, Oliver~L Haimson, and Danielle Lottridge.
\newblock How to do better with gender on surveys: a guide for hci researchers.
\newblock \emph{interactions}, 26\penalty0 (4):\penalty0 62--65, 2019.

\bibitem[S{\"u}nderhauf et~al.(2018)S{\"u}nderhauf, Brock, Scheirer, Hadsell,
  Fox, Leitner, Upcroft, Abbeel, Burgard, Milford,
  et~al.]{sunderhauf2018limits}
Niko S{\"u}nderhauf, Oliver Brock, Walter Scheirer, Raia Hadsell, Dieter Fox,
  J{\"u}rgen Leitner, Ben Upcroft, Pieter Abbeel, Wolfram Burgard, Michael
  Milford, et~al.
\newblock The limits and potentials of deep learning for robotics.
\newblock \emph{The International Journal of Robotics Research}, 37\penalty0
  (4-5):\penalty0 405--420, 2018.

\bibitem[Thieme et~al.(2020)Thieme, Belgrave, and Doherty]{datasheet3}
Anja Thieme, Danielle Belgrave, and Gavin Doherty.
\newblock Machine learning in mental health: A systematic review of the hci
  literature to support the development of effective and implementable ml
  systems.
\newblock \emph{ACM Transactions on Computer-Human Interaction (TOCHI)},
  27\penalty0 (5):\penalty0 1--53, 2020.

\bibitem[Torralba et~al.(2020)Torralba, Fergus, and
  Freeman.]{tinyimages_retraction}
A.~Torralba, R.~Fergus, and B.~Freeman.
\newblock {Tiny Images Dataset Retraction}.
\newblock \url{https://groups.csail.mit.edu/vision/TinyImages/}, 2020.
\newblock Accessed: 2020-9-20.

\bibitem[Torralba and Efros(2011)]{torralba2011unbiased}
Antonio Torralba and Alexei~A Efros.
\newblock Unbiased look at dataset bias.
\newblock In \emph{CVPR 2011}, pages 1521--1528. IEEE, 2011.

\bibitem[Torralba et~al.(2008)Torralba, Fergus, and Freeman]{torralba200880}
Antonio Torralba, Rob Fergus, and William~T Freeman.
\newblock 80 million tiny images: A large data set for nonparametric object and
  scene recognition.
\newblock \emph{IEEE transactions on pattern analysis and machine
  intelligence}, 30\penalty0 (11):\penalty0 1958--1970, 2008.

\bibitem[Wang and Kosinski(2018)]{wang2018deep}
Yilun Wang and Michal Kosinski.
\newblock Deep neural networks are more accurate than humans at detecting
  sexual orientation from facial images.
\newblock \emph{Journal of personality and social psychology}, 114\penalty0
  (2):\penalty0 246, 2018.

\bibitem[White et~al.(2020)White, Nolen, and Savani]{nas-local-search}
Colin White, Sam Nolen, and Yash Savani.
\newblock Local search is state of the art for nas benchmarks.
\newblock \emph{arXiv preprint arXiv:2005.02960}, 2020.

\bibitem[Wu and Zhang(2016)]{wu2016automated}
Xiaolin Wu and Xi~Zhang.
\newblock Automated inference on criminality using face images.
\newblock \emph{arXiv preprint arXiv:1611.04135}, pages 4038--4052, 2016.

\bibitem[Xie et~al.(2012)Xie, Zhang, You, Zhang, and Qu]{gender-skin1}
Jin Xie, Lei Zhang, Jane You, David Zhang, and Xiaofeng Qu.
\newblock A study of hand back skin texture patterns for personal
  identification and gender classification.
\newblock \emph{Sensors}, 12\penalty0 (7):\penalty0 8691--8709, 2012.

\bibitem[Xie et~al.(2019)Xie, Kirillov, Girshick, and He]{rand-wire}
Saining Xie, Alexander Kirillov, Ross Girshick, and Kaiming He.
\newblock Exploring randomly wired neural networks for image recognition.
\newblock In \emph{Proceedings of the IEEE International Conference on Computer
  Vision}, pages 1284--1293, 2019.

\bibitem[Youssef et~al.(2013)Youssef, Ibrahim, and Abbott]{gender-writing2}
Amira~E Youssef, Ahmed~S Ibrahim, and A~Lynn Abbott.
\newblock Automated gender identification for arabic and english handwriting.
\newblock In \emph{5th International Conference on Imaging for Crime Detection
  and Prevention (ICDP 2013)}, pages 1--6. IET, 2013.

\bibitem[Yu et~al.(2009)Yu, Tan, Huang, Jia, and Wu]{gender-gait1}
Shiqi Yu, Tieniu Tan, Kaiqi Huang, Kui Jia, and Xinyu Wu.
\newblock A study on gait-based gender classification.
\newblock \emph{IEEE Transactions on image processing}, 18\penalty0
  (8):\penalty0 1905--1910, 2009.

\bibitem[Zetter(2020)]{UL}
Kim Zetter.
\newblock A famed hacker is grading thousands of programs --- and may
  revolutionize software in the process.
\newblock The Intercept, \url{https://rb.gy/igtlzz}, 2020.
\newblock Accessed: 2020-9-20.

\bibitem[Zhang(2020)]{cropping}
Amy~X. Zhang.
\newblock \url{https://twitter.com/amyxzh/status/1307505876396158976?s=20},
  2020.
\newblock Accessed: 2020-9-19.

\end{thebibliography}
\end{document}